\documentclass[10pt]{article}

\usepackage[preprint]{acl}

\usepackage{times}
\usepackage{latexsym}
\usepackage[T1]{fontenc}
\usepackage[utf8]{inputenc}
\usepackage{microtype}
\usepackage{inconsolata}
\usepackage{graphicx}
\usepackage{multirow}
\usepackage{tabularx}
\usepackage{amsmath}

\usepackage{amssymb}
\usepackage{mathtools}
\usepackage{makecell}
\usepackage[table]{xcolor}
\usepackage{array}
\usepackage{appendix}
\usepackage{listings} 
\usepackage{xcolor}
\usepackage{relsize}

\title{How to Evaluate Medical AI}

\author{
  \textbf{Ilia Kopanichuk\textsuperscript{1, 2}},
  \textbf{Petr Anokhin\textsuperscript{1, 6}},
  \textbf{Vladimir Shaposhnikov\textsuperscript{1, 3}},
  \textbf{Vladimir Makharev\textsuperscript{1, 4}},
\\
  \textbf{Ekaterina Tsapieva\textsuperscript{5}},
  \textbf{Iaroslav Bespalov\textsuperscript{1}},
  \textbf{Dmitry V. Dylov\textsuperscript{1,3}},
  \textbf{Ivan Oseledets\textsuperscript{1,3}}
\\
\\
  \textsuperscript{1}AIRI, Moscow, Russia \\
  \textsuperscript{2}Moscow Institute of Physics and Technology, Dolgoprudny, Russia \\
  \textsuperscript{3}Skokovo Institute of Science and Technology, Moscow, Russia \\
  \textsuperscript{4}Innopolis University, Innopolis, Russia \\
  \textsuperscript{5}SberHealth, Moscow, Russia \\
  \textsuperscript{6}Lomonosov Moscow State University, Moscow, Russia \\
  \smaller[2]{
  \textbf{Correspondence:} kopanichuk@airi.net 
  }
\\
}

\begin{document}

\maketitle
~\\
\begin{abstract}
The integration of artificial intelligence (AI) into medical diagnostic workflows requires robust and consistent evaluation methods to ensure reliability, clinical relevance, and the inherent variability in expert judgments. Traditional metrics like precision and recall often fail to account for the inherent variability in expert judgments, leading to inconsistent assessments of AI performance. Inter-rater agreement statistics like Cohen's Kappa are more reliable but they lack interpretability. We introduce Relative Precision and Recall of Algorithmic Diagnostics (RPAD and RRAD) - a new evaluation metrics that compare AI outputs against multiple expert opinions rather than a single reference. By normalizing performance against inter-expert disagreement, these metrics provide a more stable and realistic measure of the quality of predicted diagnosis.

In addition to the comprehensive analysis of diagnostic quality measures, our study contains a very important side result. Our evaluation methodology allows us to avoid selecting diagnoses from a limited list when evaluating a given case. Instead, both the models being tested and the examiners verifying them arrive at a free-form diagnosis. In this automated methodology for establishing the identity of free-form clinical diagnoses, a remarkable 98\% accuracy becomes attainable. 

We evaluate our approach using 360 medical dialogues, comparing multiple large language models (LLMs) against a panel of physicians. Large-scale study shows that top-performing models, such as DeepSeek-V3, achieve consistency on par with or exceeding expert consensus. Moreover, we demonstrate that expert judgments exhibit significant variability - often greater than that between AI and the humans. This finding underscores the limitations of any absolute metrics and supports the  need to adopt relative metrics in medical AI.  

\end{abstract}
~\\
~\\
\section{Introduction} 
The rapid advancement of artificial intelligence in healthcare has opened new frontiers in medical diagnostics, promising faster, more accurate, and more accessible disease detection \cite{topol2019high, yu2018artificial}. As AI systems increasingly demonstrate their capability to analyze medical imaging, laboratory results, and patient data, their potential to revolutionize diagnostic processes has garnered significant attention from healthcare providers, researchers, and industry stakeholders \cite{rajpurkar2022ai, mckinney2020international, liu2019comparison}. Recent studies suggest that AI-powered diagnostic tools can match or even exceed human performance in specific areas, such as detecting certain types of cancer in radiological images or identifying pathological patterns in tissue samples.

However, the implementation of AI diagnostics faces several critical challenges that demand careful examination. Questions about the reliability, reproducibility, and generalizability of AI diagnostic systems remain paramount concerns in the medical community \cite{nagendran2020ai, he2019practical, roberts2021common}. The "black box" nature of many AI algorithms, combined with potential biases in training data and the variability in real-world clinical settings, raises important issues about the trustworthiness and clinical validity of these systems \cite{chen2021treating, obermeyer2019dissecting, gianfrancesco2018potential}. Furthermore, the integration of AI diagnostics into existing healthcare workflows presents practical challenges related to regulatory compliance, liability considerations, and the need for standardized validation protocols \cite{wiens2019no, panch2019inconvenient}.

The quality assessment of AI diagnostic systems represents a complex multifaceted problem that encompasses technical performance, clinical utility, and ethical considerations \cite{liu2019comparison, kelly2019key, mongan2020claim}. While numerous studies have demonstrated promising results in controlled settings, there is a pressing need for comprehensive evaluation frameworks that can assess these systems' performance across diverse patient populations and clinical environments . Understanding the limitations, failure modes, and potential biases of AI diagnostics is crucial for their safe and effective deployment in healthcare settings, where decisions can have profound implications for patient outcomes \cite{van2021deep, cabitza2017unintended}.

\section{Related Work}

The application of artificial intelligence to medical diagnostics has evolved significantly, with each stage of development marked by key advancements and the emergence of new challenges. The history of diagnostic systems provides important context for understanding the current focus on robust quality metrics, as these metrics are now critical to ensuring the reliability and safety of AI in healthcare.

\subsection{Evolution of Medical Diagnostic Systems}

Early diagnostic systems, such as the rule-based expert systems Mycin and DXplain, pioneered the use of AI to support clinical decision-making \cite{mycin, DXplain}. These systems relied on hand-crafted rules and knowledge bases, allowing them to provide interpretable recommendations. However, their static nature and inability to learn from new data limited their adaptability, particularly in the face of complex and varied patient presentations.

The subsequent era introduced task-specific AI models driven by machine learning techniques. Platforms like Babylon Health and Ada Health demonstrated the potential for data-driven approaches to triage and symptom assessment \cite{babylong, adahealth}. Although these models offered improvements in diagnostic speed and scalability, they often relied on narrow training datasets and single-turn interactions, which restricted their ability to engage in more comprehensive diagnostic reasoning.

The advent of large language models (LLMs) marked a turning point in medical AI. Systems such as AMIE, DISC-MedLLM, MEDAGENTS and e.t.c have integrated structured multi-turn dialogue processing, advanced natural language understanding, and expansive medical knowledge bases \cite{DISC_model, MedAgents, polaris, health-llm, amie_model, Cdd_model}.By emulating more complex diagnostic pathways and incorporating domain-specific medical data, these LLMs have set a new benchmark for what AI-driven diagnostic systems can achieve. However, they also exposed persistent challenges, including the difficulty of maintaining consistent reasoning across multi-turn dialogues and struggles with the reliability and fairness of diagnostic outputs. Introducing metrics that provide a stable and clinically meaningful assessment can help address these issues by aligning system performance evaluations with the expectations of medical professionals.

\subsection{The Importance of Quality Metrics}

As diagnostic systems have become more sophisticated, the role of quality metrics in evaluating their performance has taken center stage. Traditional accuracy measures, while informative, often fail to capture the complexity of real-world diagnostic tasks \cite{paper_survey_basic, paper_survey_advanced}. More nuanced metrics are needed to assess not only how often a system gets the correct answer, but also how well it handles ambiguous cases, rare conditions, and diverse patient populations.

Recent research emphasizes the importance of metrics that reflect clinical relevance and safety\cite{evaluating_llms_as_clinical_agents, pmc_llms_scoping_review, arxiv_llms_healthcare_review, jamanetwork_review}. For example, studies have proposed using metrics aligned with physician judgments or derived from simulation-based evaluations that mimic real-world clinical workflows \cite{evaluating_llms_as_clinical_agents, ehr_critical}. These approaches help ensure that AI systems are evaluated in contexts that closely mirror the environments in which they will be used. Additionally, stratified metrics—such as those that measure performance across different demographic groups—are becoming essential tools for identifying and mitigating biases \cite{paper_survey_metrics, llm_coding2}.

Another critical aspect is the assessment of reliability and robustness \cite{evaluating_llms_as_clinical_agents, paper_survey_advanced}. Metrics that track performance consistency across varied clinical settings and diverse patient demographics provide deeper insight into a system’s strengths and weaknesses. Over time, these quality indicators guide iterative refinements to ensure safer deployments in clinical settings and better alignment with clinical expectations.

\subsection{The Path to Stable Metrics}
% Жесть пафосно конечно
Given these challenges, we introduce a novel evaluation metric designed to align closely with expert clinical judgments. This metric provides a more stable and meaningful view of model performance by accounting for clinical relevance and reliability. Through rigorous testing in simulation-based and real-world clinical scenarios, we demonstrate that our metric better captures the nuances of medical diagnostics, offering a clearer path to understanding and improving AI diagnostic systems.

\begin{figure*}[ht]
    \centering
    \includegraphics[width=0.9\textwidth]{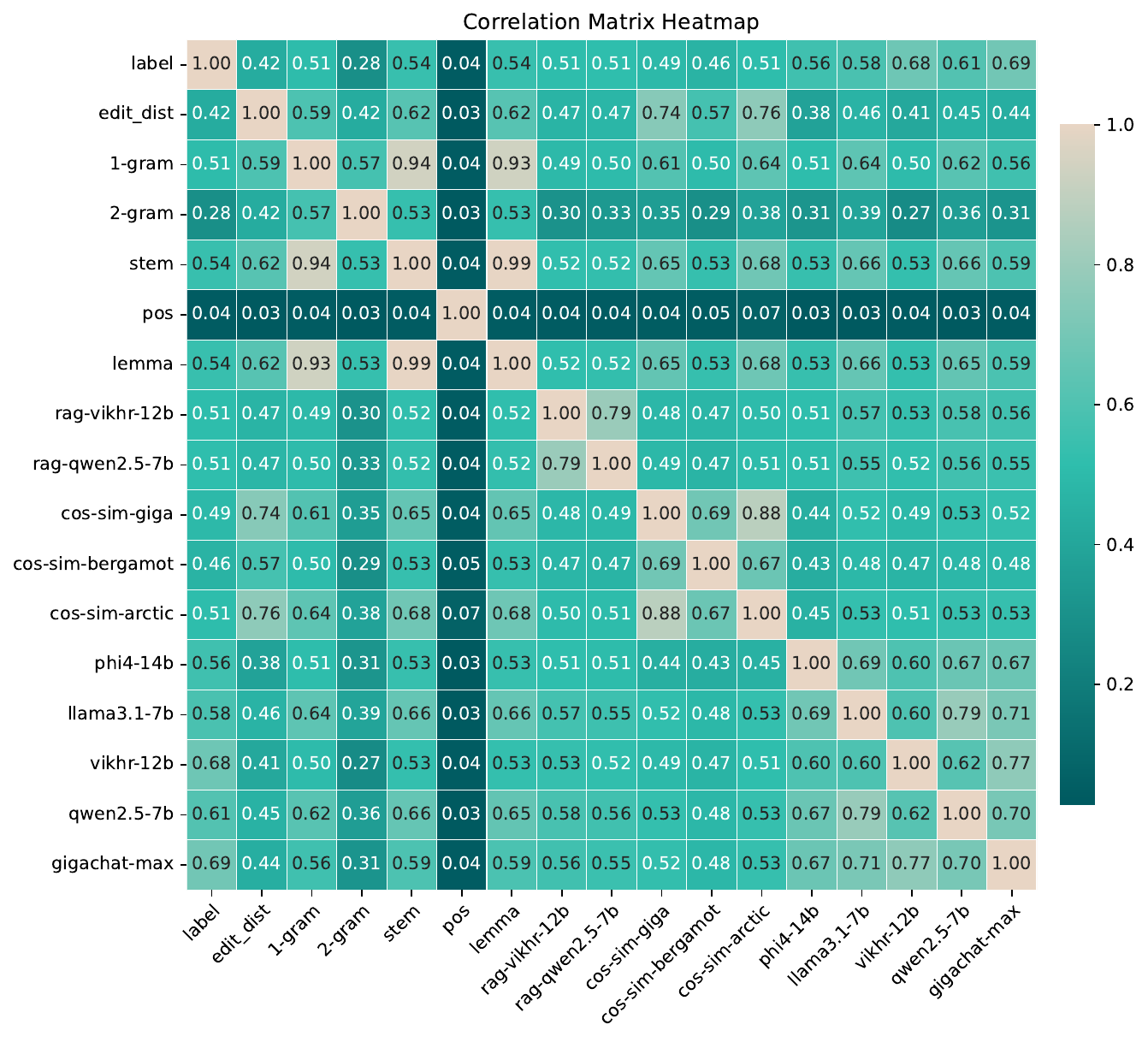}
    \caption{Meta-model (the match function $M$) features correlation matrix.}
    \label{fig:meta-modeling-train-corr}
\end{figure*}

\section{Methods} \label{exp}

The result of the algorithm's work is a list (\textit{i.e.} bag) of diagnostic hypotheses (hereinafter referred to as diagnoses). In order to evaluate the quality of the system, several problems must be solved. The first problem is the selection of evaluation metrics for the system. The fact is that the task of making a diagnosis from the text of physician's questions and patient's answers is difficult even for real human expert. Different experts may have different opinions. Consequently, it is impossible to evaluate the model based on the decisions of a single expert. In the Metrics section we propose quantitative metrics for evaluating algorithmic diagnostics: the relative precision of the algorithmic diagnostics RPAD and the relative recall of the algorithmic diagnostics RRAD. The second problem is to understand from two text strings containing a diagnosis whether they contain the same diagnosis or different ones. This problem is solved by selecting the matching function in the corresponding section.

\subsection{Metrics}

To proceed with setting up an experiment to evaluate the quality of the system, we must first define what the system is. To do this, we need to determine what will be fed into the input, what will be output from the system, and how we will be able to verify whether the output corresponds to the expected one. The input of the diagnostic system $\mathfrak{C} = \{C_1, ..., C_n\}, n \in \mathbb{N}$ is a set of medical documents $n$ $C$. $C$ is a text string that contains a chat between a patient and a doctor. Diagnosis system output $\mathfrak{D} =[\![ D_1, ..., D_n]\!] , n \in \mathbb{N}$ is a bag of $n$ bags $D = [\![ d_1, ..., d_k ]\!], k \in \mathbb{N}$, where $d$ is a text string that contains a diagnosis. The expectations of system performance will be shaped by the set of experts $\mathfrak{E} = \{E_1, ..., E_z\}, z \in \mathbb{N}$, where expert $E$ is a function $E:\mathfrak{C} \mapsto \mathfrak{D}$. The algorithm is a function $A:\mathfrak{C} \mapsto \mathfrak{D}$.  The algorithm $A$ is not in a set of experts: $\chi_{\mathfrak{E}}(A)= \emptyset$. Then we can define the pairwise precision $P$ and recall $R$ (\textit{i.e.} metrics) between the algorithm and the expert as 
\begin{equation}
    P_{AE}@k = \frac{\sum_{i=1}^{n}\mu(D^A_i, D^E_i)}{nk^2}
\end{equation}
\begin{equation}
    R_{AE}@k = \frac{\sum_{i=1}^{n}\chi(D^A_i, D^E_i)}{n}
\end{equation}
where $D_i^A = A(C_i)$, $D_i^E = E(C_i)$,  $\mu$ is a multiplicity function, and $\chi$ is a characteristic function. The pairwise metrics between two experts $E$ and $E'$  are determined similarly as $P_{EE'}@k$ and $R_{EE'}@k$. The evaluation of $\mu$ and $\chi$ is described below in the Match functions section.\\
Then the optimistic relative metrics of the algorithmic diagnosis are defined as the ratio of the maximum relative pairwise metrics of the algorithm compared to each expert to the minimum relative metrics of the expert compared to each expert that is not itself:

{\smaller[2]
\begin{equation}
    P_{opt}@k = \frac{\max{P_{A\mathfrak{E}}@k}}{\min{P_{\mathfrak{E}\mathfrak{E}}@k}} = \frac{\max\{P_{AE_1}@k,...,P_{AE_z}@k\}}{\min\{P_{E_1E_2}@k,...,P_{E_{z-1}E_z}@k\}}
\end{equation}

\begin{equation}
    R_{opt}@k =\frac{\max{R_{A\mathfrak{E}}@k}}{\min{R_{\mathfrak{E}\mathfrak{E}}@k}} = \frac{\max\{R_{AE_1}@k,...,R_{AE_z}@k\}}{\min\{R_{E_1E_2}@k,...,R_{E_{z-1}E_z}@k\}}
\end{equation}
}
And the averaged relative metrics are defined as
{\smaller[2]
\begin{equation}
    P_{avg}@k = \frac{\overline{P_{A\mathfrak{E}}@k}}{\overline{P_{\mathfrak{E}\mathfrak{E}}@k}}=\frac{\frac{1}{z}\sum_{i=1}^{z}P_{AE_i}@k}{\frac{2}{z(z-1)}\sum_{i=1}^{z-1}\sum_{j=i+1}^{z}P_{E_iE_j}@k}
\end{equation}

\begin{equation}
    R_{avg}@k = \frac{\overline{R_{A\mathfrak{E}}@k}}{\overline{R_{\mathfrak{E}\mathfrak{E}}@k}} = \frac{\frac{1}{z}\sum_{i=1}^{z}R_{AE_i}@k}{\frac{2}{z(z-1)}\sum_{i=1}^{z-1}\sum_{j=i+1}^{z}R_{E_iE_j}@k}
\end{equation}
}
Then the relative precision of the algorithmic diagnostics RPAD and the relative recall of the algorithmic diagnostics RRAD are defined as follows:
{\smaller[1]
\begin{equation}
RPAD@k = \frac{(1-H)\cdot\max{P_{A\mathfrak{E}}@k}+H\cdot\overline{P_{A\mathfrak{E}}@k}}{(1-H)\cdot\min{P_{\mathfrak{E}\mathfrak{E}}@k}+H\cdot\overline{P_{\mathfrak{E}\mathfrak{E}}@k}}
\end{equation}
\begin{equation}
RRAD@k = \frac{(1-H)\cdot\max{R_{A\mathfrak{E}}@k}+H\cdot\overline{R_{A\mathfrak{E}}@k}}{(1-H)\cdot\min{R_{\mathfrak{E}\mathfrak{E}}@k}+H\cdot\overline{R_{\mathfrak{E}\mathfrak{E}}@k}}
\end{equation}
}
where $H \in [0,1]$ is a hardness parameter. These metrics derived above have the following meaning: a value of any of them below 1.0 means that the agreement between the experts is higher than the agreement between the experts and the algorithm and vice versa.
The hardness parameter allows one to vary the score depending on the task. At $H=0$, the relative metrics take an optimistic form from eqs. 3 and 4, and at $H=1$ an average form from eqs. 5 and 6. Optimistic score is binary in nature: its value greater than one means that the model is worthy of being among the experts because it differs from the average expert less than the most deviant expert. In contrast, the average score is quantitative -- a value greater than one indicates that the model responses is closer to the average expert than the experts are to each other. \\

\subsection{Match function}

\begin{table}[htbp]
    \centering
        \caption{\label{tab:MDS} Examples of how the diagnosis match function $M$ works. $d^A$ is the algorithm response, $d^E$ is the expert response.}
        \begin{tabular}{|c|c|c|}
         \hline
                $\mathbf{d^A}$ & $\mathbf{d^E}$ & \textbf{M} \\ \hline
                \makecell{pelvic  \\ inflammatory \\ disease} & \makecell{interstitial \\ cystitis } & 1  \\ \hline
                \makecell{acute \\ pancreatitis} & \makecell{chronic \\ pancreatitis \\ exacerbation} & 0\\ \hline
                \makecell{rotavirus \\ gastroenteritis} & \makecell{acute \\ intestinal \\ infection \\ gastroenteritis}  & 1  \\ \hline
                \makecell{foodborne \\ toxicosis} & \makecell{viral infection}  & 0  \\ \hline
                \makecell{arrhythmia} & \makecell{functional \\ heart rhythm \\ disorders syndrome}  & 1  \\ \hline
    \end{tabular}
\end{table}

A key part of the experiment is to calculate the values of functions $\chi$ and $\mu$ in eqs. 1 and 2. As the algorithm $d^A$ and the expert $d^E$ responses are text strings, how can we find which of pairs $(d^A,d^E)$ do match or not? Strings do not have to be identical to contain the same diagnosis, see Table \ref{tab:MDS}. \\ 
To solve this problem, let $\mathfrak{F}$ be a set of pairs of text strings. Then any match function should be $M:\mathfrak{F}\mapsto \mathbb{B}$. Now we can calculate the multiplicity $\mu$ and the characteristic $\chi$ functions for diagnoses in case of $|D^A|~=~|D^E|~=~k$ as long as we have a proper definition for a match function:

\begin{equation}
   \mu(D^A, D^E) = \sum_{p=1}^{k}\sum_{q=1}^{k} M(d^A_p, d^E_q)
\end{equation}

\begin{equation}
    \chi(D^A, D^E) = 
    \begin{dcases}
        1; & \mu(D^A, D^E) > 0 \\
        0; &   \mu(D^A, D^E) = 0
    \end{dcases}    
\end{equation}

%To evaluate the alignment between AI-generated and expert diagnoses, a reliable method for comparing diagnostic statements is essential. This requires a robust diagnosis matching function capable of determining whether two diagnostic expressions, potentially varying in form but equivalent in meaning, refer to the same underlying condition.

To this end, we implemented a supervised meta-modeling approach based on comprehensive feature engineering. The resulting model, used to define the match function $M$, predicts whether a pair of diagnoses $(d^A, d^E)$ should be considered a match.

\subsubsection*{Feature Construction and Preprocessing}

The match function was trained on labeled diagnosis pairs through a multi-stage pipeline. First, we applied preprocessing to normalize the data, including lowercasing, punctuation removal, abbreviation expansion, and domain-specific rule corrections. After preprocessing, the train and test sets comprised 4,833 and 1,469 samples, respectively. The train and test sets were curated by 3 medical experts: The final value of the label was decided by a vote of each of them. Another expert additionally reviewed the test dataset in accordance with the criteria developed by the same experts together. The criteria developed by experts to assess the similarity of diagnoses are shown in Appendix \ref{apx:prompt-direct-llm}. An example of a dataset can be seen in Table \ref{tab:MDS}.

To train the meta-model, we extracted 17 features for each pair, grouped into four categories:

\begin{enumerate}
    \item \textbf{Direct LLM prompting:} Language models were directly instructed to assess diagnosis similarity (see \ref{apx:prompt-direct-llm} for the prompt). The same criteria developed by expert physicians that guided them in manually marking up the dataset were used to write the prompt. Five models (Llama-3.1-70B-Instruct, Qwen2.5-7B-Instruct, GigaChat-Max, Vikhr-Nemo-12B-Instruct, Phi-4) %какие модели 
    where selected out of 20 
    %сколько моделей тестировали
    by the highest F1 score on the test dataset. 
    %так подбирали?
    Each model’s output contributed a binary feature indicating whether it considered the pair a match.

    \item \textbf{LLM with RAG and ICD dictionary:} We used a Retrieval-Augmented Generation (RAG) system indexed on 20,000+ ICD entries. Each diagnosis was matched to ICD code using an LLM-guided retrieval process. We generated an embedding for each entry in the ICD-10 library using the BERGAMOT \cite{sakhovskiy-etal-2024-biomedical} model,
    % linking_model_bergamot_biosyn_ep19 (from other work)
    stored in FAISS 
    %какую эмбеддинг модель использовали. faiss или просто косинус?
    vector store \cite{douze2024faiss} for Euclidean distance-based similarity search. For each input diagnosis, a direct match was first attempted. If no direct mapping was found, the top 15 most similar ICD-10 entries were retrieved using vector similarity.
    Two LLMs (Qwen2.5-7B-Instruct and Vikhr-Nemo-12B-Instruct) were then employed separately in different experiments to identify the most relevant ICD entry from the retrieved candidates (see \ref{apx:prompt-rag-llm} for the prompt). The input diagnosis was subsequently assigned the ICD code corresponding to the model's selected entry.
    % Two LLMs (Qwen2.5-7B-Instruct, Vikhr-Nemo-12B-Instruct) 
    %какая ллм
    % were then used (one LLM for each experiment) to select the most appropriate entry from the retrieved candidates (see Appendix for prompts) and the input diagnosis was assigned the ICD code of the selected entry.
    Two diagnoses were considered a match if their ICD codes shared the same first three characters, contributing a binary feature to the meta-model.
    
    \item \textbf{Text embedding similarity:} We used three embedding models: Giga, BERGAMOT \cite{sakhovskiy-etal-2024-biomedical}, and Arctic-Embed \cite{merrick2024arcticembedscalableefficientaccurate} to compute cosine similarity between diagnosis vectors. These scores served as continuous features.
    
    \item \textbf{Linguistic similarity ratios:} For the fourth group of features, we computed linguistic similarity ratios between pairs of diagnoses. For each diagnosis, we extracted sets of unigrams, bigrams, stems, lemmas, and part-of-speech (POS) tags. Similarity ratios were calculated as the size of the intersection between corresponding sets divided by the minimum size of the sets. In addition, we computed the normalized Levenshtein distance between the diagnosis strings to capture character-level differences.
\end{enumerate}

\subsubsection*{Meta-Model Training and Evaluation}

Model selection and training were performed using the AutoGluon AutoML framework \cite{erickson2020autogluontabularrobustaccurateautoml}. Optimization targeted average precision on cross-validation folds. Training was constrained to a maximum of one hour with automatic ensembling (stacking) enabled. The resulting best-performing model was a 
% LightGBM classifier.
bagged ensemble of LightGBM classifier as a base model.
% (\texttt{LightGBM\_r196\_BAG\_L1}).

\subsubsection*{Feature Importance and Selection}
\begin{figure}[h!]
    \centering
    \includegraphics[width=0.45\textwidth]{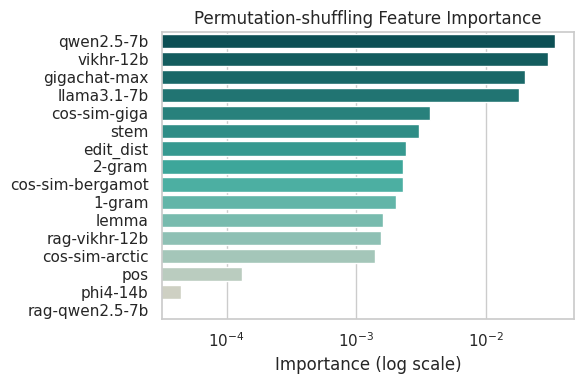}
    \caption{Meta-model (the match function $M$) features importance by permutation-shuffling.}
    \label{fig:meta-modeling-test-feature-importance-log}
\end{figure}

\begin{figure}[b!]
    \centering
    \includegraphics[width=0.45\textwidth]{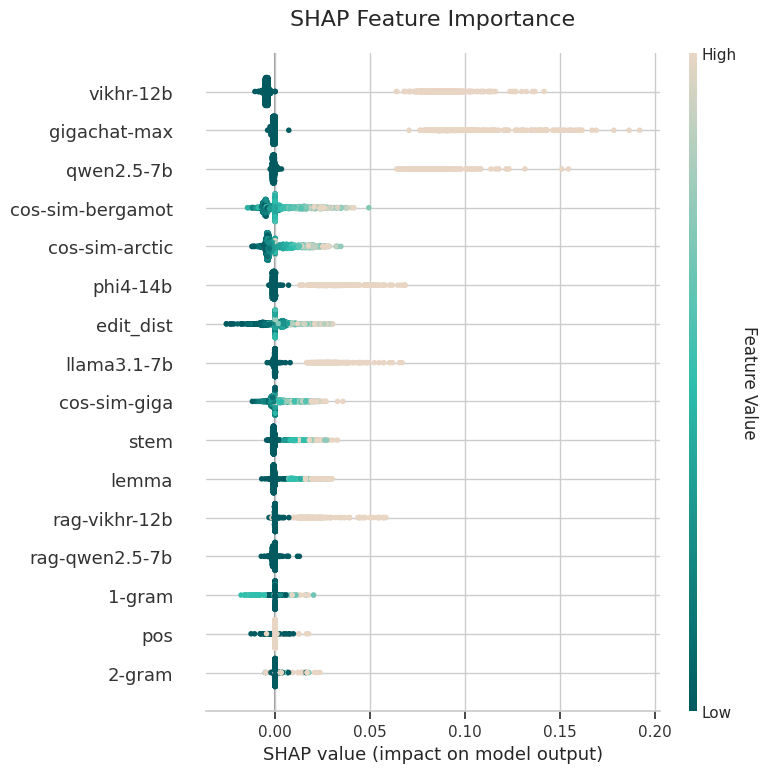}
    \caption{Meta-model (the match function $M$) features importance by SHAP values.}
    \label{fig:meta-modeling-test-shap-200}
\end{figure}

We initially filtered features using correlation analysis (see Fig. \ref{fig:meta-modeling-train-corr}), excluding low-informative candidates such as 2-gram and Part-of-speech ratios, but observed no performance gain. Final selection was based on permutation-shuffling feature importance provided by AutoGluon (see Fig. \ref{fig:meta-modeling-test-feature-importance-log}), which indicated all features contributed non-negatively. Additionaly, the SHAP values analysis (see \ref{fig:meta-modeling-test-shap-200}) validated the importance of each feature by providing a game theory-based measure. Hence, the full feature set was retained for deployment.

Since we introduce custom match function, they have their own quality metrics. The problem of quality assessment of such a function reduces to a binary classification problem, and we use standard metrics for it: precision, recall, $F_1$-measure, and accuracy. Table~\ref{tab:MFM} reports the evaluation metrics for the final match function $M$, which underpins the computation of $\mu$ and $\chi$ in our diagnostic comparison experiments.\\

\begin{table}[htbp]
    \centering
        \caption{\label{tab:MFM} Match functions metrics}\begin{tabular}{|c|c|c|c|c|c|}
         \hline
               \textbf{Precision}& \textbf{Recall}& \boldmath{$F_1$}  &\textbf{Accuracy}& \textbf{Support} \\ \hline
               0.91 &  0.90  & 0.91  & 0.98 & 1469 \\ \hline
    \end{tabular}
\end{table}

\section{Results and Discussion}

\subsection{Data}
The raw dataset is a collection of $n=360$ chats. The first half of the dataset consists of chats from two experts played out a pre-determined scenario. One of them knew the disease and their symptoms and was the patient. The other tried to define the disease by asking questions as the physician. The second half of the dataset consists of the chats between the actor playing a patient and LLM playing a physician. Ground truth labels for the dataset were marked by $z=7$ experts, all of them are resident physicians. Each expert marked up all chats with $k_{max}=3$ diagnoses. Also did the algorithm. \\

\subsection{LLM comparison}
Based on the data from experts, we calculated pairwise and relative metrics for the diagnostic algorithm using different LLMs. For comparison, we selected the LLMs that show the best metrics on the MERA dataset \cite{fenogenova-etal-2024-mera} (see Table \ref{tab:quality_diagnosis}).

\begin{table*}[htbp]
    \small
    \caption{\label{tab:quality_diagnosis}{ Relative quality of the algorithmic diagnostics for different LLMs }}
    \centering
    \begin{tabular}{|c|c|c|c|c|}
    \hline
   \textbf{LLM} & \boldmath{$P_{avg}@1$} & \boldmath{$P_{opt}@1$} & \boldmath{$R_{avg}@3$} & \boldmath{$R_{opt}@3$} \\ \hline
    GigaChat-Max                      & 1.09 & + & 1.12 & + \\ \hline
%    GigaChat-Pro                      & 0.91 & + & 0.97 & + \\ \hline
    Qwen-72B                          & 1.00 & + & 1.06 & + \\ \hline
    \textbf{DeepSeek-V3}                       & \textbf{1.15} & + & \textbf{1.16} & \textbf{+} \\ \hline
    DeepSeek-R1                       & 0.88 & \textbf{+} & 1.11 & + \\ \hline
    Dist-Qwen32B         & 0.88 & + & 0.91 & - \\ \hline
    GPT4o                             & 1.13 & + & 1.13 & + \\ \hline
    Mistral-Large                     & 1.01 & + & 1.09 & + \\ \hline
    Llama-405B                        & 0.88 & + & 0.92 & - \\ \hline
    \end{tabular}
\end{table*}

Table \ref{tab:quality_diagnosis} shows the relative diagnostics metrics for different LLMs. The proposed metrics enable a comprehensive and objective assessment of the diagnostic capabilities of language models. Their interpretation provides not only a quantitative measure of performance but also distinguishes between the model’s best-case potential (via optimistic metrics) and its overall practical reliability (via averaged metrics). Taken together, these indicators establish DeepSeek-V3 as the most reliable candidate for integration into clinical decision support systems. However, the difference in its quality with GigaChat-Max and GPT4o is not significant enough to be meaningful. Dist-Qwen32B and Llama-405B provides the worst metrics among studied models. A value of $R_{opt}@3 < 1$ means that these two models cannot even be considered as candidates for inclusion in the pool of diagnostic algorithms.

\begin{figure}[htbp]
    \centering
    \includegraphics[width=0.45\textwidth]{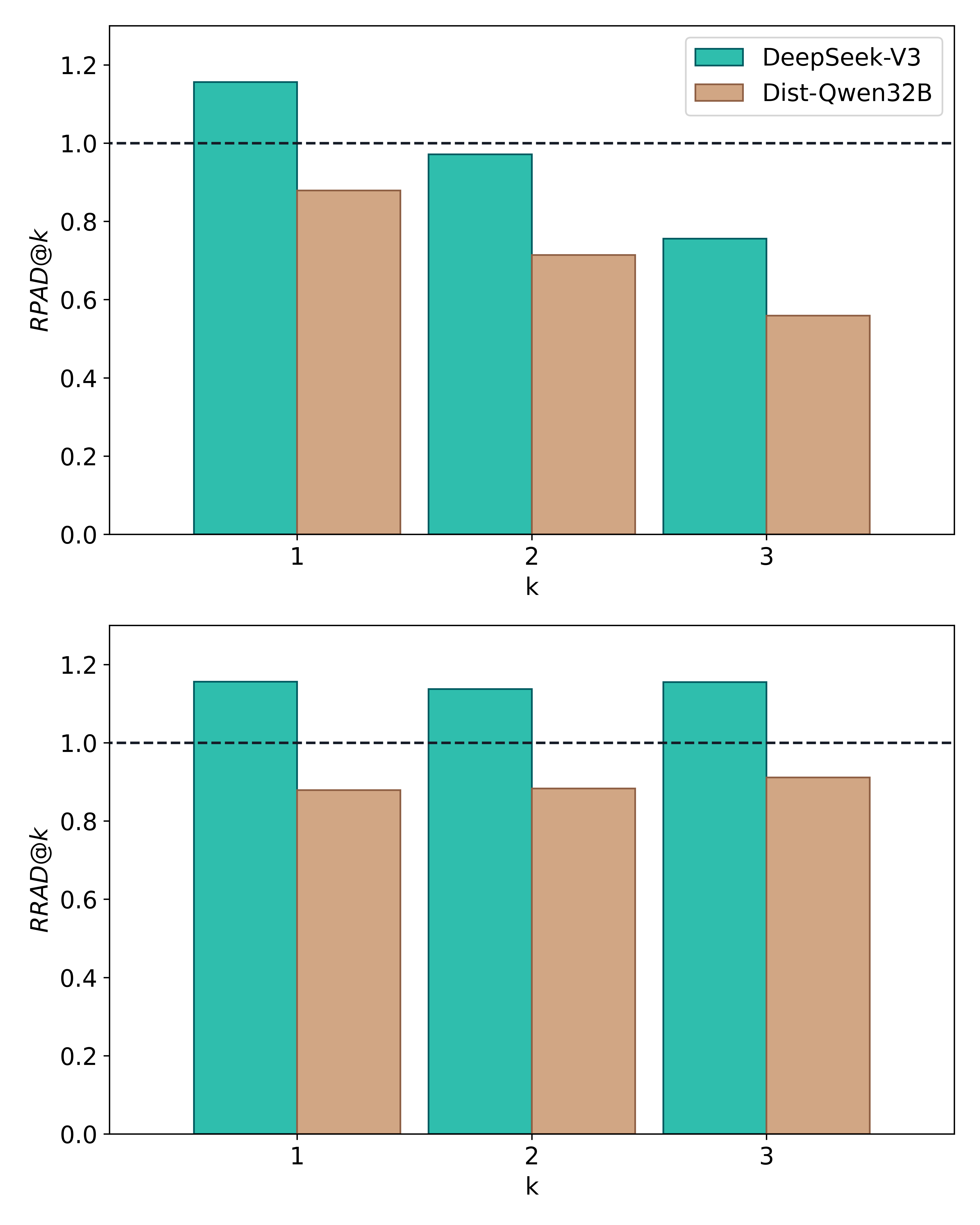}
    \caption{Relative precision and recall of the algorithmic diagnostics for the best and the worst models from Table \ref{tab:quality_diagnosis}.}
    \label{fig:topk_diag}
\end{figure}

Figure \ref{fig:topk_diag} illustrates the dependence of relative precision RPAD@k and relative recall RRAD@k values on the number of returned diagnoses at the highest possible hardness $H=1$. \textbf{DeepSeek-V3} model consistently demonstrates the highest values across the entire range of $k$ for both metrics. Furthermore, the plot confirms that DeepSeek-V3 reliably outperforms experts in terms of relative recall, making it a preferred choice for the development of clinical decision support systems. Both plots feature a dashed line corresponding to the value of $1.0$ --- this serves as an important visual reference: values above this threshold indicate that the model’s agreement with experts is at least on par with, or even exceeds, the level of agreement among the experts themselves. DeepSeek-V3 crosses this line in several instances, especially in RRAD@k, highlighting its alignment with - and at times superiority to - inter-expert consensus.

\subsection{Variance of expert responses}

\begin{figure}[htbp]
    \centering
    \includegraphics[width=0.45\textwidth]{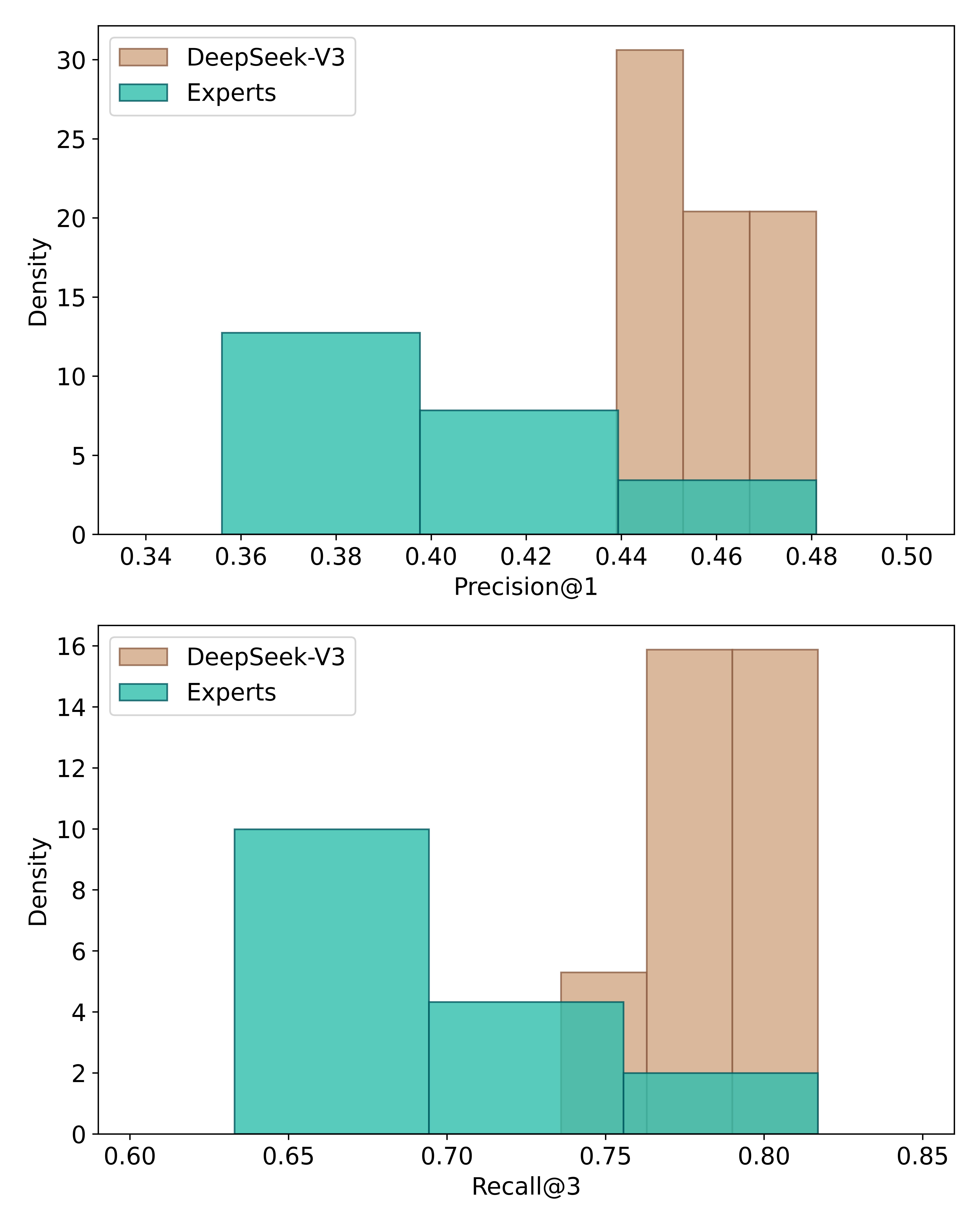}
    \caption{Distributions of the algorithm-expert and expert-expert pairwise precision and recall}
    \label{fig:distribution} 
\end{figure}

\begin{figure}[htbp]
    \centering
    \includegraphics[width=0.45\textwidth]{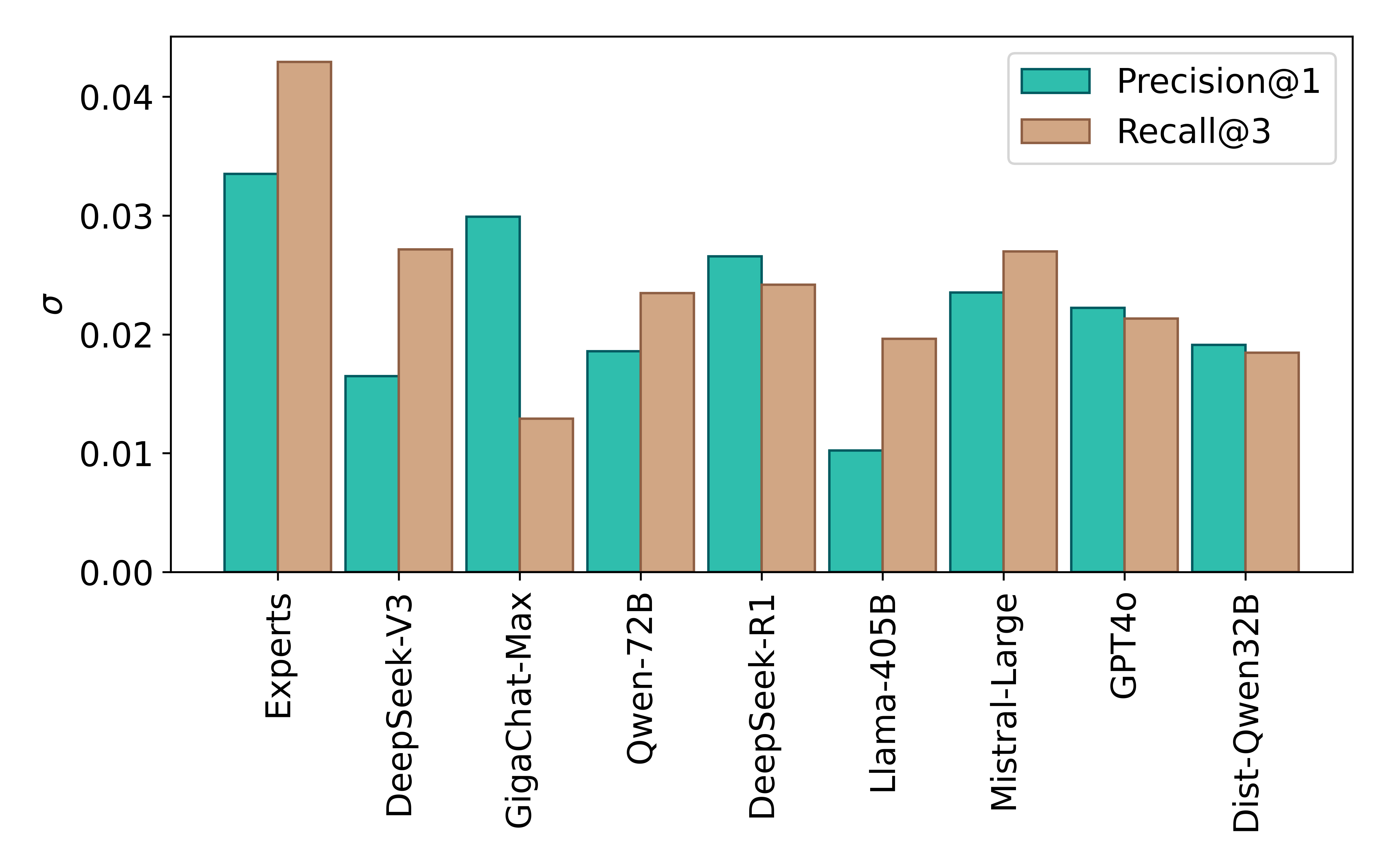}
    \caption{Unbiased estimation of standard deviation of pairwise precision and recall for each LLM and expert}
    \label{fig:variance} 
\end{figure}

To test the original hypothesis of variability in expert opinions stated at the beginning of Section \ref{exp}, we computed the distributions of precision and recall of expert and algorithm responses. An example of this distribution is demonstrated in Figure \ref{fig:distribution} for the DeepSeek-V3 model. It is noticeable that the model responses are clustered more tightly than the experts' responses relative to each other. To make it more convincing, we calculated the unbiased estimate of the standard deviation of the precision and recall for all considered LLMs relative to the experts (Eq. 11) and the average unbiased estimate of the standard deviation of the experts' metrics relative to each other  (Eq. 12). Figure \ref{fig:variance} shows that variance of experts' responses is significantly higher than variance for each model's responses, sometimes by several times. This is why we prefer to use the relative metrics RPAD and RRAD when evaluating the quality of algorithmic diagnostics. The differences in expert opinions are so high that absolute metrics can differ by a factor of several in the selection of one expert or another, and it will be very difficult to prefer one opinion or another.

\begin{equation}
    \small
   \sigma^{A}_{R} = \sqrt{\frac{1}{z-1}\sum^{z}_{i=1} (R_{AE_i}@k - \overline{R_{A\mathfrak{E}}@k})^2}
\end{equation}

\begin{equation}
    \small
   \overline{\sigma^{\mathfrak{E}}_{R}} = \frac{1}{z}\sum^{z}_{i=1} \sqrt{ \frac{1}{z-2} \sum^{z}_{j \neq i} (R_{E_iE_j}@k - \overline{R_{E_i\mathfrak{E}}@k})^2}
\end{equation}

% \end{table}
\subsection{Cohen's Kappa}

\begin{figure}[htbp]
    \centering
    \includegraphics[width=0.45\textwidth]{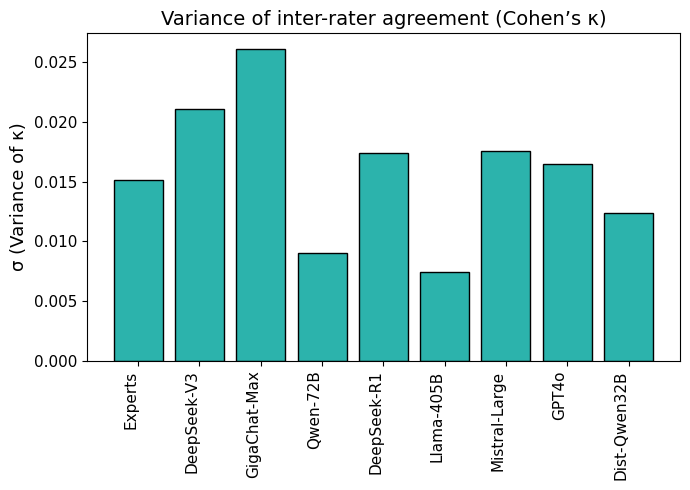}
    \caption{Unbiased estimation of standard deviation of pairwise agreement $\kappa$ for each LLM and expert}
    \label{fig:kappa_var} 
\end{figure}

For completeness, we plotted the distributions of Cohen’s $\kappa$ for the same cases, computing unbiased variance estimates \ref{fig:kappa_var} of $\kappa$ for each model and for every pair of experts (Eqs.~13–14). The $\kappa$  distributions for pairs consisting of DeepSeek-V3 and an expert completely overlap with those for pairs of experts. This overlap implies that the metric lacks the resolution needed to reliably distinguish between models. This lack of resolution originates from several intrinsic limitations of the $\kappa$ statistic, specifically:

\begin{itemize}
  \item \textbf{Top‑1 only.} Standard $\kappa$ considers only the first‐ranked diagnosis, ignoring the relevance of other hypotheses, which makes it ill‐suited to multi‑choice settings.
  \item \textbf{Top‑$k$ bias.} Generalizing $\kappa$ to top‑$k$ metrics triggers the "inflated agreement" effect \cite{Carpentier2017-sc}, where response aggregation artificially boosts consensus estimates while obscuring meaningful  variation.
    \item \textbf{Clinical granularity.} $\kappa$ treats every label as strictly distinct; clinically synonymous or closely related diagnoses are counted as full disagreement, which makes the statistic poorly adapted to medical taxonomies with overlapping or hierarchical categories.
\end{itemize}

In contrast, the \textit{RPAD} and \textit{RRAD} distributions computed from the same unbiased estimates but normalized by inter‑expert variance—remain well separated and robust to the choice of reference expert and number of hypotheses. Therefore, under high variability in expert opinions, the \emph{relative metrics} (RPAD, RRAD) provide a more reliable and interpretable means of comparing algorithmic diagnostic performance.

\begin{equation}
   \sigma^{A}_{\kappa} = \sqrt{\frac{1}{z-1}\sum^{z}_{i=1} (\kappa_{AE_i} - \overline{\kappa_{A\mathfrak{E}}})^2}
\end{equation}

\begin{equation}
   \overline{\sigma^{\mathfrak{E}}_{\kappa}} = \frac{1}{z}\sum^{z}_{i=1} \sqrt{ \frac{1}{z-2} \sum^{z}_{j \neq i} (\kappa_{E_iE_j} - \overline{\kappa_{E_i\mathfrak{E}}})^2}
\end{equation}

% \end{table}
\section{Conclusions} 

This study highlights a critical challenge in evaluating AI diagnostic systems: the high variability in expert judgments. Our analysis revealed that disagreements among human experts are often greater than the differences between AI models and expert consensus. This variability makes traditional absolute metrics unreliable for assessing AI performance, as results can vary significantly depending on which expert's opinion is used as the benchmark.

To address this issue, we introduced relative metrics (RPAD and RRAD), which compare AI outputs against the range of expert opinions rather than a single reference. These metrics provide a more stable and realistic evaluation, addressing the variability in human diagnoses and providing a clearer measure of reliability.  

Since medical terms can vary in phrasing while conveying the same meaning, a simple text comparison is insufficient. Instead, we developed a supervised meta-model that combines multiple features including direct LLM assessments, text embeddings, and linguistic similarity to make this decision. The model was trained on expert-annotated data to ensure clinical relevance. This approach allows flexible yet accurate matching, enabling fair comparisons between AI and human diagnoses in our evaluation framework.

Our findings suggest that AI models can achieve consistency exceeding that of human experts. However, the high variance in expert judgments underscores the need for standardized evaluation frameworks that consider multiple perspectives. Future work should explore ways to reduce expert disagreement, such as clearer diagnostic guidelines or consensus-building methods, to further improve AI assessment in healthcare. Testing on 360 cases showed that models like DeepSeek-V3 perform well, sometimes matching or even exceeding expert agreement. However, differences between top models were small, highlighting the need for careful selection in clinical settings.  

The findings emphasize that AI diagnostics should be assessed not just for accuracy but also for consistency and alignment with expert opinions. Recognizing and accounting for expert variability is essential for fair and meaningful evaluations of AI diagnostic tools. Our approach offers a practical solution to this challenge, supporting better integration into healthcare. Future work should expand testing to more diverse medical scenarios and improve evaluation methods further.

\bibliography{main}

\newpage

\appendix
\lstdefinestyle{prompt}{
  backgroundcolor=\color{gray!10},
  basicstyle=\ttfamily\small,
  frame=single,
  breaklines=true,
  keepspaces=true,
  breakindent=0pt
}

\section{LLM Prompts Used}
The original prompts were in Russian. Below are their translations to English.

\subsection{Direct prompt for LLM}
\label{apx:prompt-direct-llm}
\begin{lstlisting}[style=prompt]
Motivation: To create a reference metric for evaluating the accuracy of free-form diagnosis formulation by an AI model.

Task: Compare the formulations of two diagnoses, one made by a doctor and one by artificial intelligence, and determine their similarity. Answer only: Yes or No.

Formulations will be considered similar if they meet one of the following six criteria:

1. If one diagnosis specifies the other.
2. If one diagnosis specifies the stage of the other.
3. If the formulation of one diagnosis is a synonym for the formulation of the other.
4. If one of the diagnoses is a laboratory indicator of the other diagnosis.
5. If the formulations are completely identical.
6. If the diagnoses are identical but contain a spelling error.

Diagnosis 1: {diag}
Diagnosis 2: {other_diag}
\end{lstlisting}

\subsection{Prompt for LLM with RAG and ICD dictionary}
\label{apx:prompt-rag-llm}

\begin{lstlisting}[style=prompt]
You will be given a reference diagnosis and a list of diagnoses from a database. Your task is to determine which diagnosis from the database most closely matches the reference diagnosis. Try to choose the diagnosis accurately, paying attention to details. Choose the diagnosis with the most word and meaning matches. You can only choose from the diagnoses in the list. Pay more attention to the diagnoses at the beginning of the list, as they are more likely to be suitable. It is better to choose a shorter diagnosis than one that contains information not present in the reference diagnosis. In your response, write only the diagnosis number and nothing else.

Reference diagnosis: {candidate}
Diagnoses list from the database: {diagnoses}
\end{lstlisting}

\end{document}